\documentclass[journal]{IEEEtran}

\usepackage{algorithm}
\usepackage{algorithmic}
\usepackage{amsmath}
\usepackage{amssymb}
\usepackage{array}
\usepackage{bm}
\usepackage{booktabs}
\usepackage{cases}
\usepackage[pdftex]{graphicx}
\usepackage{indentfirst}
\usepackage{mathrsfs}
\usepackage{multirow}
\usepackage{url}
\usepackage{xcolor}
\usepackage{makecell}
\usepackage{pifont}
\usepackage{subfig}
\usepackage{bbding}

\begin{document}
%
\title{Video Is Graph: Structured Graph Module for Video Action Recognition}
%
\author{Rong-Chang Li,
        Xiao-Jun~Wu,
        and~ Tianyang~Xu
\thanks{R.-C. Li, X.-J. Wu, and T. Xu are with the School of Artificial Intelligence and Computer Science, Wuxi, P.R. China. (e-mail: li\_rongchang@stu.jiangnan.edu.cn; wu\_xiaojun@jiangnan.edu.cn; tianyang.xu@jiangnan.edu.cn)}}
%
%
%
%
%
\maketitle
\sloppy
\begin{abstract}
In the field of action recognition, video clips are always arranged as ordered frame sequences. 
Therefore, previous studies rely on conducting communication between adjacent frames to obtain spatio-temporal features.
They require a redundant stacking mechanism to achieve long-range global perception. 
In this paper, we first propose to transform a video sequence into a frame graph to obtain direct long-term dependencies. 
Then, to preserve sequential information during transformation, we devise a structured graph module (SGM). 
SGM divides the neighbours of each node into several temporal regions so that it can extract global structural information and sequence features simultaneously, with a slight computation increase. 
Extensive experiments are performed on widely-used typical datasets, such as Something-Something V1 \& V2, Diving48, Kinetics-400, UCF101, and HMDB51. 
The results show that SGM can achieve competitive or SOTA results with less computation overhead.
\end{abstract}
%
%
%
\IEEEpeerreviewmaketitle
\section{Introduction}
The video action recognition task that aims to identify the action category of a video clip plays a significant role in video surveillance, human-computer interaction, autonomous driving industries, etc. 
However, video data brings more challenges as it is the projection of the spatio-temporal real world.  
Despite the recent rapid development in the community of deep learning \cite{feichtenhofer2020x3d, li2020tea, feichtenhofer2019slowfast, sudhakaran2020gate, yang2020temporal, wang2020video, wu2019spatial, weng2020temporal}, extracting discriminant spatio-temporal features from the original video data is still a very significant theme with huge research value, and feature extraction \cite{wu2004new, zheng2006nearest, zheng2006reformative} plays a crucial part in action recognition, as well as in image fusion \cite{luo2016novel, luo2017image, li2020nestfuse}  and many other computer vision tasks \cite{li2011no, li2013no, chen2018new, sun2011quantum, sun2019effective, wang2003initial}.
Without considering supplementary clues such as optical flow, current end-to-end methods can be divided into two categories: spatio-temporal joint learning methods based on 3D convolution \cite{tran2015learning, feichtenhofer2020x3d, feichtenhofer2019slowfast, qiu2017learning, tran2018closer, carreira2017quo, hara2018can, xie2018rethinking,9351626} and spatio-temporal separation learning methods based on additional hand-craft temporal feature learning modules \cite{zhou2018temporal, li2020tea, sudhakaran2020gate, weng2020temporal, liu2020teinet, lin2019tsm, zhu2019approximated}. 

\begin{figure}
\begin{center}
\includegraphics[width=1\linewidth]{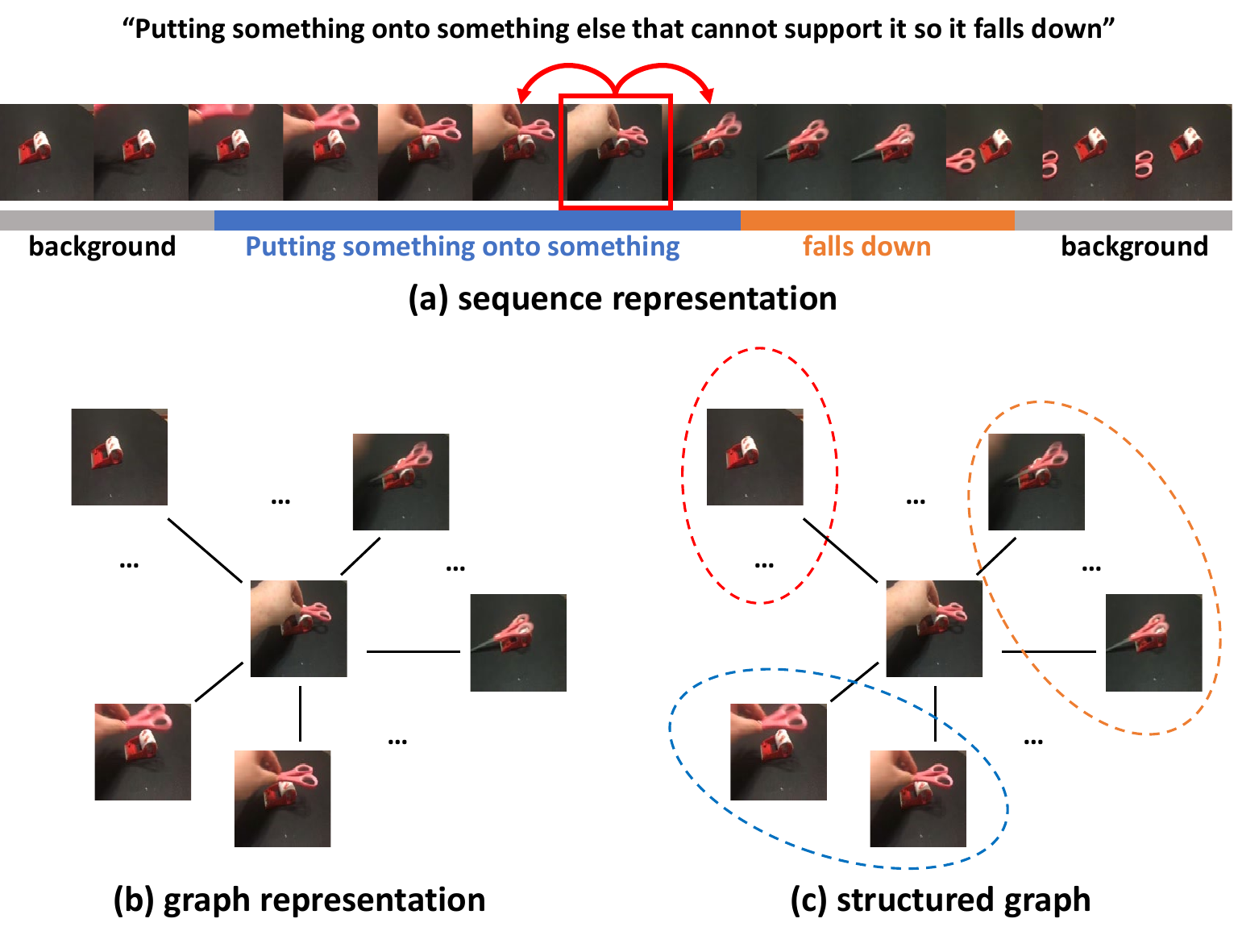}
\end{center}
\caption{We use a specific frame (in the {\color{red}{red}} box) to compare the three different organization forms of a video clip. \textbf{(a)} is the current popular sequential representation, in which each frame can only communicate with adjacent frames. \textbf{(b)} is a graph representation, in which it can communicate with all frames, but the sequence information is destroyed. \textbf{(c)} is a structured graph representation, at this situation it communicates with different temporal regions, so that global structural information and sequence features can be obtained simultaneously.}
\label{motivation}
\end{figure}

In general, their mechanism is very similar, i.e., focusing on local temporal feature extraction and increasing network depth or stacking temporal modules to obtain long-term dependencies. 
Though some hidden long-term patterns can be captured, the inherent defects of this mechanism make it complicated to summarize global information. 
First, for a specific temporal module, its task is to extract local information in the current stage. 
Therefore, useful spatial information for the next or further stages is easily neglected.
Second, at different network depths, spatial details are continuously distorted due to convolution and pooling operations. 
When the temporal receptive field expands with depth, it is uncertain whether any meaningful spatial information is still preserved or not. 
In addition, this mechanism only considers the temporal cues within local Windows, thus lacking the perception of the global temporal structure.

In this paper, we reorganize the frames in a video clip to form an interconnected frame graph. 
Based on the discussion above, we assume all frames are interlinked by learnable edges so that the video sequence is transformed into a complete graph. 
We adopt the graph convolution method to share the information of each frame along the edges, to realize the extraction of long-range dependencies.
However, after being converted to a graph, each frame in the video clip treat its neighbour frame equally, so the natural sequence information of the video clips is seriously damaged. 
To tackle this problem, we devise the Structured Graph Module (SGM). 
SGM is designed to group neighbour nodes according to temporal interval and temporal direction thus dividing these nodes into different temporal regions.
In Fig.~\ref{motivation}, we compared the three organization forms of a video clip. 

By grouping, the original complete graph is divided into several allopatric sub-graphs and each subgraph contains the relationships between each node and its specific temporal region under the specific grouping principle. 
Since subgraphs are gained according to the original sequence attributes of the video clip, inference results from different subgraphs contain corresponding temporal patterns.
Therefore, SGM can extract global structure information and temporal sequence features. 
In addition, unlike the multi-head attention mechanism repeatedly with varied parameters on an identical structure, SGM separately transmit information. 
In this way, SGM can force the blind information transmission process to be more concentrated but without superfluous operations.

To gain sufficient spatio-temporal features, we insert SGM with the InceptionV3 \cite{szegedy2016rethinking} network to construct an SGN network.
We evaluate it on short-interval motion-focused datasets (Something-Something V1$\&$V2 \cite{goyal2017something, mahdisoltani2018effectiveness}), long-interval motion-focused datasets (Diving48 \cite{li2018resound}), and scene-focused datasets (Kinetics-400 \cite{carreira2017quo}, UCF101 \cite{2012UCF101}, HMDB51 \cite{2011HMDB}). 
Our SGN can achieve on par with or better than the latest competitions on these datasets, with a marginal increase in computation (1.08$\times$ as many as InceptionV3).

The main contributions of this paper are summarized as follows:
\begin{itemize}
\item 
We propose to reorganize the frames in a video clip and transform it into graph structures to capture long-term dependencies.
\item 
We propose a novel Structured Graph Module (SGM), which groups the neighbours of nodes according to the temporal prior information so that sufficient sequence information can be saved in the transformation process from sequence representation to graph representation.
\item 
We construct a novel network SGN by inserting SGM to inception-v3 standard blocks. Due to the innovative SGM, the SGN can extract various spatio-temporal features and perceive global structural information. Finally, our SGN achieves SOTA performance on various datasets with a marginal increase in computation. 
\end{itemize}

\section{Related work}
\textbf{Video action recognition.} How to extract spatio-temporal features has mirrored advances in video action recognition. 
Early works \cite{simonyan2014two, feichtenhofer2016convolutional,7847353} respectively learned appearance and motion information via inputting RGB or optical flow (or its alternative) to 2D-CNNs. 
However, calculating and storing optical flow are extremely costly.
To this end, extracting temporal information such as motion from raw RGB frames becomes an important research topic.

3D-CNNs \cite{tran2015learning, hara2018can} are the dilatant of 2D CNNs, which can jointly learn spatio-temporal features with equal treatments to all dimensions. 
Later studies proposed a series of creative works based on 3D CNNs. 
To alleviate the optimization difficulty, I3D \cite{carreira2017quo} inflated pretrained 2D kernels to 3D. 
P3D \cite{qiu2017learning} and R(2+1)D \cite{tran2018closer} decompose 3D convolution into temporal (1D) and spatial (2D) convolution. 
To reduce the computation overhead, S3D \cite{xie2018rethinking} and ECO \cite{2018ECO} use different convolution types in the different stages of the network. 
To develop more capabilities, slow-fast \cite{feichtenhofer2019slowfast} adopts two branches that respectively focus on appearance and motion information. 
TPN \cite{yang2020temporal} utilizes the output at each stage to capture action instances at various tempos. 
However, 3D-CNNs view the temporal dimension as a simple expansion of traditional 2D spatial dimensions, thus lacking the consideration of the inherent data distribution between spatial and temporal relevance. 
As a result, the 3D framework needs a dense temporal sampling rate and longer input sequence to gain satisfying performance. 

\begin{figure*}[t]
\begin{center}
\includegraphics[width=1\linewidth]{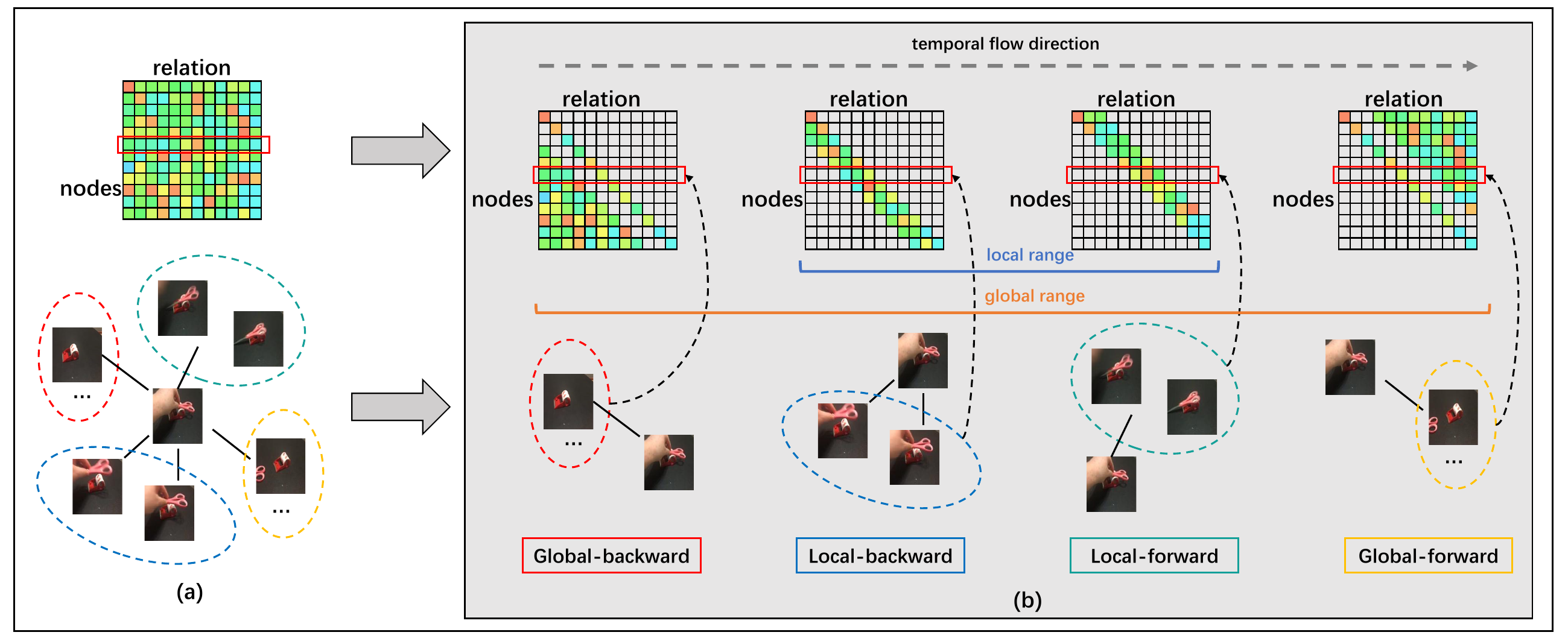}
\end{center}
\caption{The group process of the structured graph. Figure (a) is the original complete graph and its corresponding adjacency matrix. Each row of the adjacency matrix represents the relationship between the node at that index position and other nodes. The coloured ellipses in Figure (a) represent regions divided by different grouping principles. The grouping principle, subregions divided and the corresponding adjacency matrix are shown in Figure (b).}
\label{sgm}
\end{figure*}

Another type is the spatio-temporal separation learning manner \cite{lin2019tsm, liu2020teinet, li2020tea, sudhakaran2020gate, 9447897}. 
The key is to add spatio-temporal modelling capabilities to the original 2D network. 
In this paradigm, TSM \cite{lin2019tsm} introduces shift operation to achieve interaction between neighbour frames. 
TEI \cite{liu2020teinet} parameterizes shift operation and uses neighbour difference to excite channels. 
TEA \cite{li2020tea} proposes two sequential modules to excite motion information and aggregate multiple temporal information. 
GSM \cite{sudhakaran2020gate} proposes a fine-grained gate to control shift operation between adjacent frames. 
These methods share the same schema that firstly models short-term temporal information and then stack layers (or operations) to expand the receptive field, which is less efficient with limited long-term patterns being preserved.

\textbf{Long-term temporal modelling.} previous studies try to capture long-term dependency at the late stage, such as using RNN \cite{2015Beyond,2017Long} or multi-scale MLPs \cite{zhou2018temporal}.
Non-local \cite{} proposes to build direct dependencies between spatio-temporal pixels. 
StNet \cite{2019StNet} designs hierarchically manner to learn local and global information. 
V4D \cite{2020V4D} adds clip-level convolutions in the later stages to aggregate long-term information. 
Recently, transformer-based methods \cite{2021Is, 2021ViViT} are also proposed to directly model long-term patterns.

\textbf{Graph methods for video action recognition} While there are considerable works in downstream video understanding tasks and skeleton-based action recognition, less attention has been paid to video action recognition.
Wang \cite{wang2018videos} firstly introduces graph to video action recognition but it relies on extracted spatio-temporal features from I3D. 
TRG \cite{zhang2020temporal} tries to insert GAT \cite{2017Graph} modules to different stages, neglecting the discrepancy between video sequences and traditional node data. 
DyReG \cite{duta2020dynamic} uses RNN to generate spatio-temporal nodes and utilize MLP to send messages along edges and finally adoptes another GRU \cite{chung2014empirical} unit to update each nodes. 
However, it filters out much more scene information, with a complicated and heavy design.

\section{Approach}
In this section, we first explain how to represent a video clip with graph (sec{\color{red}~\ref{3_1}}). 
Then we introduce the structured graph module (SGM) that divides neighbours into different temporal regions according to temporal prior information (sec{\color{red}~\ref{3_2}}). 
Finally, we discuss how to integrate our proposed module (SGM) to the existing 2D network to gain multi-scale spatio-temporal modelling ability (sec{\color{red}~\ref{3_3}}).

\subsection{General graph representation of video clips.}
\label{3_1}


Here we present a general graph structure representation for video clips. 
Given a video, we first uniformly divide a video into $T$ segments and then select a frame from each segment to form the video's sparse representation, $F=\{f_0,...f_{T-1}\}$. 
Here $F$ denotes the set of input frames and $f_i$ is the $i$-th frame (Index $i$ also implies the temporal order). 
Next, we express the corresponding graph representation of $F$ as $G=(V,E)$ ($V, E$ denote the node set and edge set of the graph). 
Different from traditional graph nodes which are separate individuals, a frame clip is essentially a sparse sampling representation of a single video. 
From this perspective, when we try to build the graph between frames from a video, we aim at extracting the relations between different temporal components of the video. 
In this way, we can assume that there is a universal graph that can describe how the different temporal components rely on each other. 
As discussed above, we designate frame as nodes of graph $G$ i.e., $V=F$ and set $E=\{e_{i,j}\}$ as learnable weights.

Once obtaining the weighted edge set $E$, we can directly gain the corresponding adjacent matrix $A \in \mathbb{R}^{T\times{T}}$. 
Then the process to aggregate and update each nodes can be formulated as:
\begin{equation}
\begin{aligned}
{y}_{i}=\operatorname{ReLU}\left(\sum_{j \in \mathcal{N}_{i}} \alpha_{i j} \mathbf{W}{x}_{j}\right)
\end{aligned}
\end{equation}
where $\alpha_{i j}$ represents the $i$-th row $j$-th column element in the adjacent matrix $A$, $ \mathbf{W}$ is originally a linear transformation's weight matrix but here it is replaced by a convolution operation and ${y}_{i}$ is the reasoning result at the $i$-th temporal point. 

\subsection{Structured Graph Module (SGM)}
\label{3_2}
After we transform a video clip into a graph structure, the connections between the different components of this video clip are established so the long-range dependencies can be captured directly. 
However, the graph structure breaks the original sequential arrangement of frames, resulting in the loss of sequential information. 
To solve this problem, we attempt to utilize the prior information of the ordered sequence to guide the information flow in the established graph structure. 
In practice, we divide the neighbours of each node into several groups by the principles related to temporal attributes, and decompose the initial complete graph into the corresponding sub-graphs. 
We consider two physical properties of video sequences, i.e., \textbf{1) Temporal direction}. A video sequence consists of multiple frames that occur in chronological order, thus the temporal direction is of much importance for capturing temporal clues such as ‘from right to
left’ or ‘from left to right’. \textbf{2) Temporal interval}. The human can perceive instantaneous motion and displacement from a small temporal window or global semantic change from connections of large temporal intervals. 
That means connections with different temporal span contain different temporal scale information.

\begin{table*}[h]
\begin{center}
\begin{tabular}{|c|c|c|c|c|c|c|c|}
\hline
\multirow{2}*{Method} & \multirow{2}*{backbone} &\multirow{2}*{Frames}&\multirow{2}*{GFLOPs}&\multicolumn{2}{|c|}{V1}&\multicolumn{2}{|c|}{V2}\\
\cline{5-8}
~&~&~&~&Top-1(\%) &Top-5(\%)&Top-1(\%) &Top-5(\%)\\
\hline\hline
TSN-RGB \cite{wang2016temporal}        & BNInception      &8  &16     & 19.5   & -  & -   & -  \\
TRN-Multiscale \cite{zhou2018temporal}     & BNInception      &8  &33     & 34.4   & -   & 48.3   & 77.6 \\
S3D-G \cite{xie2018rethinking}             & Inception        &64  &71.38     & 48.2  &78.7   & -  & - \\
GSM \cite{sudhakaran2020gate}         & InceptionV3        &16  &53.7     & 50.6&-  & -   & - \\
\hline 
TSM \cite{lin2019tsm}               & ResNet-50         &8  &33     & 45.6 &74.2  & -   & - \\
TSM \cite{lin2019tsm}              & ResNet-50         &16  &65    & 47.2 &77.1  & 63.4   & 88.5 \\
TSM \cite{lin2019tsm}               & ResNet-50         &8+16  &98    &49.7 &78.5  & -   & - \\
\hline 
TEINet \cite{liu2020teinet}             & ResNet-50         &8  &33     & 47.4 &-  & 61.3   & - \\
TEINet \cite{liu2020teinet}             & ResNet-50         &16  &66    & 49.9 &-  & 62.1   & - \\
TEINet \cite{liu2020teinet}             & ResNet-50         &(8+16)$\times$30  &99$\times$30     & 52.5 &-  & 66.5   & - \\
\hline 
TEA \cite{li2020tea}            & ResNet-50         &16  &70     & 51.9&80.3  & -   & - \\
TAM \cite{fan2019more}            & ResNet-50         &16$\times$2  &47.7     & 48.4 &78.8  & 61.7   & 88.1 \\
\hline 
ECO \cite{2018ECO}          & BNIncep+R18       &92  &267     & 46.4&-  & -   & - \\
I3D \cite{carreira2017quo}         & ResNet-50        &32$\times$2  &306     & 41.6&72.2  & -   & - \\
GST \cite{luo2019grouped}        & ResNet-50        &16  &59     & 48.6&77.9  & 62.6   & 87.9 \\
STM \cite{jiang2019stm}        & ResNet-50        &16$\times$30  &67$\times$30    & 50.7&80.4  & 64.2   & 89.8 \\
V4D \cite{2020V4D}         & ResNet-50        &8$\times$4  &167.6     & 50.4&-  & -   & - \\
SmallBigNet \cite{li2020smallbignet}        & ResNet-50        &8+16  &157     & 50.4&80.5  & 63.3   & 88.8 \\
\hline 
SGN(ours)           & InceptionV3        &8  &\textbf{25.4}     & \textbf{48.9} &\textbf{77.2} & \textbf{61.6} &\textbf{87.5}\\
SGN(ours)           & InceptionV3        &16  &\textbf{50.8}     & \textbf{51.2} &\textbf{78.9} & \textbf{63.1} &\textbf{88.5}\\
SGN(ours)           & InceptionV3        &(16+8)$\times$2$\times$3  &\textbf{76.2}$\times$6     & \textbf{54.9} &\textbf{82.4} & \textbf{67.1} &\textbf{90.9} \\
\hline 
\end{tabular}
\end{center}
\caption{Comparison with state-of-the-art methods on Something-Something V1 $\&$ V2. Our proposed SGN can achieve better performance with less computation.} 
\label{something_sota}
\end{table*}

As described above, we have transformed a video frame sequence $F=\{f_1,...f_T\}$ into a graph representation $G=(V,E)$. 
To make the grouping concise, we divide the edge set $E$ into several allopatric edge set:
\begin{equation}
\begin{aligned}
{E} = {E_0} \cup {...} \cup {E_{n-1}}\\ 
\forall i \neq j,  { E_i} \cap {E_j} = \varnothing
\end{aligned}
\end{equation}
Here $n$ is the number of group principles. 
We adopt two sequence attributes, temporal direction and interval, as the basis for grouping. 
First, we divide the neighbors of nodes into local and global nodes according to temporal interval:
\begin{equation}
\begin{aligned}
{e_{i,j}} \in 
 \left\{\begin{matrix}
{E_0,} & {if} &{\left|i - j\right|\leqslant \tau}\\ 
{E_1,} & {if} &{\left|i - j\right| > \tau}\\ 
\end{matrix}\right.
\end{aligned}
\end{equation}
Intuitively, $E_0$ contains edges of small temporal span while $E_1$ contains edges of large temporal span, so these two sets respectively contain local or global information. 
Then, according to the order direction, we further classify the set as:
\begin{equation}
\begin{aligned}
{e_{i,j}} \in 
 \left\{\begin{matrix}
{E_0,} & {if} &{ i - j < -\tau}\\ 
 {E_1,} & {if} &{ -\tau\leqslant i - j \leqslant 0}\\  
{E_2,} & {if} &{ 0\leqslant i - j \leqslant \tau}\\  
{E_3,} & {if} &{ \tau< i - j }\\  
\end{matrix}\right.
\end{aligned}
\label{f4}
\end{equation}
As $i$ and $j$ respectively represent the frame indexes connected by $e_{i,j}$, the values of $i$ and $j$ also denote the temporal order. 
Thus the local and global regions are divided into two sub-set (forward or backwards) according to the relative temporal order.
The corresponding adjacent matrix $A_k$ to each edge set $E_k$ is construct as:
\begin{equation}
\begin{aligned}
{\alpha_{i,j}^k} =
 \left\{\begin{matrix}
{\alpha_{i,j}} & {if} &{{e_{i,j}} \in {E_k}}\\ 
{0} & {if} &{{e_{i,j}} \notin  {E_k}}\\ 
\end{matrix}\right.
\end{aligned}
\end{equation}
where $\alpha_{i j}$ represents the $i$-th row $j$-th column element in adjacent matrix $A$, and $\alpha_{i j}^k$ represents the $i$-th row $j$-th column element in the adjacent matrix $A^k$.

\begin{figure}
\begin{center}
\includegraphics[width=1\linewidth]{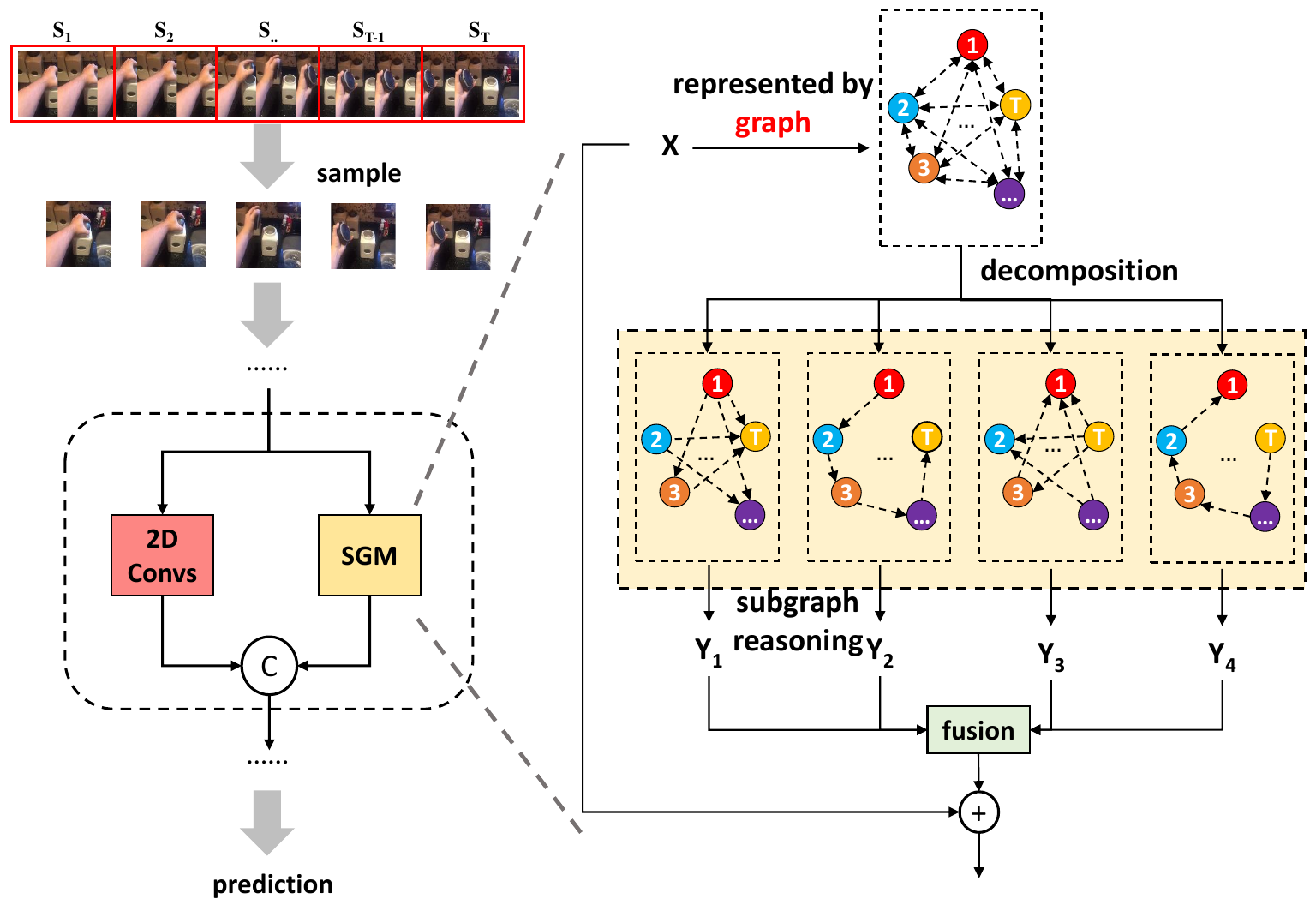}
\end{center}
\caption{Overall network architecture of SGN. SGM and the 2D convolutions from InceptionV3 together constitute a module capable of extracting various spatiotemporal features. The forward propagation process of SGM is on the right.}
\label{network}
\end{figure}

As shown in Fig.~\ref{sgm} (a), the relationship between nodes (or frames) in the established graph structure is reflected in the adjacent matrix. 
Each row of the adjacent matrix represents the relationships between the node at the corresponding index position and the nodes at other positions. 
Therefore, grouping the neighbours of each node corresponds to dividing each row of the adjacent matrix. 
Since different neighbour grouping policies extract different subgraphs from the original graph structure, there are consistent one-to-one match relations among the grouping policies, the subgraphs and the adjacency matrixes. 
Fig.~\ref{sgm} (b) reflects the corresponding relationship among them.

After gaining the divided sub-graphs, we first conduct reasoning on each sub-graph to obtain specific temporal features such as local-forward, global-backwards, etc. 
Then all the sub-reasoning results would be gathered by a fusion operation. The whole process is formulated as:
\begin{equation}
\begin{aligned}
{y}_{i}=\operatorname{Fuse}\left({y}_{i}^{0}, {y}_{i}^{1} ... {y}_{i}^{n-1}\right)
\end{aligned}
\end{equation}
where ${y}_{i}^k$ represents the reasoning result of the $i$-th sub-graph at the $i$-th time point and $\operatorname{Fuse()}$ could be either $concat$ + $convolution$ or directly the $sum$ operation.
As shown in (3), each sub-graph reasoning result is formulated as:
\begin{equation}
\begin{aligned}
{y}_{i}^k=\operatorname{ReLU}\left(\sum_{j \in \mathcal{N}_{i}} \alpha_{i j}^k \mathbf{W}^k{x}_{j}\right)
\end{aligned}
\end{equation}
where $\alpha_{i j}^k$ represents the $i$-th row $j$-th column element in the adjacent matrix $A^k$, $ \mathbf{W^k}$ is the corresponding convolution operation. Fig.~\ref{network} shows this process.

\begin{table}
\begin{center}
\begin{tabular}{|c|l|c|c|}
\hline
&Paradigm & graph structure & Top-1(\%)\\
\hline\hline
&2D backbone     & -         & 18.47    \\
\hline
\textcircled{\small{1}}&Inductive        & full     & 38.32  \\  
\textcircled{\small{2}}&Transductive     & full     & \textbf{47.47}  \\
\textcircled{\small{3}}&Inductive        & l\&g     & 41.62  \\
\textcircled{\small{4}}&Transductive     & l\&g     & \textbf{47.88}  \\
\textcircled{\small{5}}&Inductive        & l\&g (directional)     & 47.54 \\ 
\textcircled{\small{6}}&Transductive     & l\&g (directional)     & \textbf{48.94}  \\
\textcircled{\small{7}}&Transductive    & full$\times${4}    & {47.44}  \\
\hline
\end{tabular}
\end{center}
\caption{Ablation study results of paradigms and graph structure. All the results show the effectiveness of graph methods. \textcircled{\small{1}} vs \textcircled{\small{2}}, or \textcircled{\small{3}} vs \textcircled{\small{4}}, or \textcircled{\small{5}} vs \textcircled{\small{6}} give the specific comparison between inductive and transductive paradigms.  \textcircled{\small{1}} vs \textcircled{\small{3}} vs \textcircled{\small{5}}, or \textcircled{\small{2}} vs \textcircled{\small{4}}, vs \textcircled{\small{6}} give the specific comparison between different graph structures. These results confirm that transductive paradigm is much better and structurally decomposing graph is very effective. } 
\label{ab1}
\end{table}

\textbf{Threshold separating local or global regions.}  
As shown in Formula ~\ref{f4}, the parameter $\tau$ divides nodes into temporal local range and temporal global region. 
We explore how the different values of $\tau$ influence the performance.
Table.~\ref{ab2} shows the results of ablation experiments. 
We found that performance degrades when thresholds are too large or too small, because improper partitioning can jumble up different discriminative temporal 
features. 
Following the result in Table.~\ref{ab2}, we finally set $\tau$ as $T/8$.

\subsection{Network architecture}
\label{3_3}
The SGM module can be easily combined with current advanced 2D networks.
Considering that SGM is designed to capture various temporal cues, we choose the InceptionV3 network as our backbone as it can effectively model multi-scale spatial patterns. 
Following the practical experience of GST \cite{luo2019grouped} and GSM \cite{sudhakaran2020gate} in designing networks, we insert the SGM module into the Inception standard modules. 
In this way, the various temporal feature from SGM and the multi-scale spatial features from other inception branches are cascaded together to form semantically spatio-temporal features. 
The overall network architecture is shown in Fig.~\ref{network}.

\section{Experiments}
We evaluate our methods on 6 challenging and typical datasets. 
We first introduce these datasets and implementation details. 
Then we show the ablation study and compare our model with other SOTA methods.

\begin{table}
\begin{center}
\begin{tabular}{|c|c|c|c|c|}
\hline 
&$\tau$=1/16   &$\tau$=1/8   & $\tau$=1/4  & $\tau$=1/2\\
\hline\hline
Top-1 ($T$=8)     & - & \textbf{48.94\%}& 48.16\% & 47.87\%   \\
Top-1 ($T$=16)     & 50.46\%  & \textbf{51.21\% }& -&-  \\
\hline
\end{tabular}
\end{center}
\caption{Ablation study results of temporal threshold. Here $T$ is the total number of frames in the input video clip.}
\label{ab2}
\end{table}

\begin{table}[h]
\begin{center}
\begin{tabular}{|c|c|c|c|}
\hline 
Fusion strategy & Param. &Flops   & Top-1(\%)\\
\hline\hline
cascade   & {27.3M}  &{28.4G}    & 48.88 \\  
addition            & \textbf{24.0M}  &\textbf{25.4G}    &\textbf{48.94}  \\
\hline
\end{tabular}
\end{center}
\caption{Ablation study results of fusion strategies.}
\label{ab3}
\end{table}

\subsection{Datasets and implementation details}
\textbf{Motion-focused datasets,} including Something-Something-V1 \cite{goyal2017something} \&V2 \cite{ mahdisoltani2018effectiveness}) and Diving48 \cite{li2018resound}. 
Something-Something is a large-scale dataset. 
As collected by performing the same actions with a different object in different scenes, Something-Something demands more temporal modelling requirements from action recognition methods. 
Something-Something has two versions. 
The first one consists of 86,017 training videos and 11,522 validation videos belonging to 174 action categories while the second contains more training videos (168,913) as well as validation videos (24,777) with the same categories. 
Besides, samples in Something-Something are of 2$ \sim $4 seconds so this dataset is focused on Short-term actions. 
Diving48 is a fine-grained video dataset of competitive diving, consisting of 18k trimmed video clips of 48 unambiguous dive sequences. 
As designed with no significant biases towards static or short-term motion representations, Diving48 is suitable to assess the ability to model long-term and fine-grained dynamics information. 
In addition, samples in Diving48 usually contains several distinct stages of diving action, so it is suitable to examine the ability to perceive the global structure of video clips. 

\textbf{Scene-focused datasets,} including UCF101, HMDB51, Kinetics-400. 
All of these three datasets are scene-focused and even a single frame would often contain enough information to predict the category. 
Kinetics-400 consists of approximately 240k training and 20k validation videos trimmed to 10 seconds from 400 human actions. 
UCF101 and HMDB51 are small datasets.
UCF101 includes 13,320 videos with 101 action classes.
HMDB51 contains 6766 videos with 51 categories. 
These two small datasets are very suitable to verify the transferability of the model.

\textbf{Implementation details.} We adopt InceptionV3 pretrained on Imagenet as our backbone. 
We randomly sample $T$ frames from a video as the input sequence. 
Then the short spatial size is resized to 256 and the final spatial size is cropped to 229$\times$229 (to match the input size of InceptionV3.) 
\textbf{During training}, we do random cropping and flipping as data augmentation. 
The network is trained using SGD with an initial learning rate ($lr$) of 0.01 and momentum of 0.9 on two GPUs. 
We use a cosine learning rate schedule to update $lr$ at each epoch. 
The total number of training epochs is set as 60 with the first 10 epochs used for gradual warm-up. 
The batch size is 32 for $T=8$ and 16 for $T=16$.
\textbf{During inference,} for efficient comparison, we just use a single clip and a centre-crop with the size of 229$\times$229 for evaluation. 
For accuracy comparison, we adopt 2 clips and 3 crops with the size of 261$\times$261 to get the final average prediction.

\begin{table}[h]
\begin{center}
\resizebox{\columnwidth}{!}{
\begin{tabular}{|c|c|c|c|c|c|}
\hline
Method             & backbone          &Frames                  &GFLOPs            &Top-1    &Top-5\\
\hline\hline
TSN-RGB \cite{wang2016temporal}           & InceptionV3      &25$\times$1$\times$10    &3.2$\times$250    &72.5         &90.2  \\
S3D-G \cite{xie2018rethinking}             & InceptionV1      &64$\times$10$\times$3    &71.4$\times$30    &74.7         &93.4  \\
TSM  \cite{lin2019tsm}              & ResNet-50        &16$\times$10$\times$3    &65$\times$30     &74.7         &91.4  \\
TEINet \cite{liu2020teinet}             & ResNet-50        &16$\times$10$\times$3    &66$\times$30     &76.2         &92.5  \\
TEA  \cite{li2020tea}               & ResNet-50        &16$\times$10$\times$3    &70$\times$30     &76.1         &92.5  \\
TAM \cite{fan2019more}                & ResNet-50        &48$\times$3$\times$3    &93.4$\times$9     &73.5         &91.2  \\
\hline 
R(2+1)D \cite{tran2018closer}           & ResNet-34        &32$\times$10$\times$1    &152$\times$10     &74.3         &91.4  \\
NL I3D \cite{wang2018non}            & ResNet-50        &128$\times$10$\times$3   &282$\times$30     &76.5         &92.6  \\
SlowFast \cite{feichtenhofer2019slowfast}           & ResNet-50        &(4+32)$\times$10$\times$1    &36.1$\times$10     &75.6    &92.1  \\
\hline 
SGM(ours)           & InceptionV3     &8$\times$3$\times$3   &\textbf{25.4$\times$9}       & \textbf{73.6} &\textbf{91.2} \\
SGM(ours)           & InceptionV3     &16$\times$3$\times$3  & \textbf{50.8$\times$9}      &\textbf{75.4}  &\textbf{92.1} \\
SGM(ours)           & InceptionV3     &(8+16)$\times$3$\times$3  &\textbf{76.2$\times$9}   & \textbf{76.2} &\textbf{92.6} \\
SGM(ours)           & InceptionV3     &(8+16+24)$\times$3$\times$3  &\textbf{152.4$\times$9}   & \textbf{77.0} &\textbf{93.0} \\
\hline 
\end{tabular}}
\end{center}
\caption{Comparison with state-of-the-art methods on Kinetics-400.}
\label{kinetics_sota}
\end{table}

\begin{table}[h]
\begin{center}
\resizebox{\columnwidth}{!}{
\begin{tabular}{|c|c|c|c|c|}
\hline
Method        & pretrained      & backbone              &UCF101    &HMDB51\\
\hline\hline
TSN \cite{wang2016temporal}           &ImageNet         & InceptionV2           &86.4\%         &53.7\%  \\
P3D \cite{qiu2017learning}          &ImageNet         & ResNet-50             &88.6\%         &-  \\
C3D \cite{tran2015learning}          &Sports-1M         & ResNet-18           &85.8\%         &54.9\%  \\
I3D \cite{carreira2017quo}         &ImageNet+Kinetics         & InceptionV2           &95.6\%         &74.8\%  \\
S3D \cite{xie2018rethinking}         &ImageNet+Kinetics         & InceptionV2           &96.8\%         &75.9\%  \\
TSM \cite{lin2019tsm}         &Kinetics          & ResNet-50           &96.0\%         &73.2\%  \\
STM \cite{jiang2019stm}          &ImageNet+Kinetics         & ResNet-50           &96.2\%         &72.2\%  \\
TEA  \cite{li2020tea}         &ImageNet+Kinetics         & ResNet-50           &96.9\%         &73.3\%  \\
\hline 
SGM(ours)    &ImageNet+Kinetics         & InceptionV3        &\textbf{95.6\%}       &\textbf{78.1\%} \\
\hline 
\end{tabular}}
\end{center}
\caption{Comparison with state-of-the-art methods on UCF101 and HMDB51.} 
\label{ucf_sota}
\end{table}

\subsection{Ablation Study}
We report the ablation experiment result on the Something-Something V1 dataset. 
All the results are referenced with an efficient set, i.e., a single clip with centre-crop. 

\textbf{Study of SGM.}
We first set up two learning paradigms to determine the weight of edges in the graph. 
In the first paradigm, we take frames as nodes and suppose that each video sample owns a unique graph structure. 
So the specific edge weights between node pairs can be determined by attention-based methods \cite{2017Graph}. 
In another paradigm, we suppose there is a universal graph that can describe how the different temporal components in a video clip rely on each other so these frames are connected with a fixed weighted graph. 
According to whether the train and test datasets share the same graph structure, we call the first inductive paradigm and the second transductive paradigm. 

For each learning paradigm, we gradually add temporal prior information to decompose the graph structure. 
So there is three different graph structures: \textbf{full} (without decomposition), \textbf{l\&g} (decomposed into local and global subgraphs), \textbf{l\&g (directional)} (decomposed into four subgraphs: local-forward, local-backward, global-forward and global-backward). 
Table.~\ref{ab1} shows the results under different settings. 
Firstly, the performance of the transductive paradigm is better in all situations, which indicates that depending on the matching degree of spatial semantics may mislead the information flow in the graph. 
We believe the main reason is that this method only considers the spatial semantic similarity of the two nodes when determining the edge weight connecting them. 
Thus it neglects other information such as the direction and the position of edges in the overall structure. 
This also illustrates that there is still a migration gap between the video field and the traditional graph model.

As Table.~\ref{ab1} shows, information of temporal interval and temporal direction provides 3.3\% and 5.92\% improvement for the inductive paradigm, 0.41\% and 1.06\% improvement for the transductive paradigm. 
Finally, temporal priors provide a startling improvement of almost 9.22\% and 1.7\% for the two paradigms, respectively. 
We also compare the results of fusing four complete graphs and find that simply increasing the number of graphs does not provide gains. 
This fully illustrates the merit of our proposed SGM.

\textbf{Fusion method.}
As for the strategy of fusing the inference results of subgraphs, we compare the operation of direct addition and convolution after cascade.
Table.~\ref{ab3} shows the computational overhead, parameters and performance under different fusion strategies. 
We finally adopted the additive strategy due to its fewer parameters, lower computation, and slightly better performance.

\subsection{State-of-the-art comparison}
\textbf{Something-Something}. In Something-Somehing datasets, Different categories of samples share some common scenes, objects. 
So the Something-Something datasets are widely used to evaluate the temporal modelling capability. 
Table.~\ref{something_sota} reports the results on the Something-Something V1 and Something-Something V2 datasets. 
In terms of the advanced methods we compared, TSN and TRN use 2D networks with late fusion.
TSM, TEI, TEA, TAM and GSM adopt 2D networks combined with the temporal module.
While other methods are recent methods using 3D modules. 
Results on both datasets consistently prove that our SGN can achieve nearly the best performance with the lowest computational overhead. 
Of the comparative approaches, our network comes closest to the framework adding temporal modules to 2D backbones. 
TSM and GSM use temporal channel shift operation to simulate temporal convolution.
TEI and TAM use depth-wise temporal convolution to parameterize shift operation and add excitation or multi-branch structure to enhance temporal information. 
TEA utilizes temporal difference excitation mechanisms and Res2Net-like structures to extend the receptive field of temporal convolution. 
In contrast, our proposed SGM does not carry out the complicated manual intervention, but only makes the model automatically optimized from the perspective of graph structure, which exceeds the previous method with the lowest cost.

\textbf{Diving48}. Different from the Something-Something dataset, Diving48 is a diving action dataset with a longer video duration and distinct stages of actions. 
We use the latest version of the annotations and report the results of 16 frames with a single clip or double clips. 
Table.~\ref{diving_sota} shows the results. 
In the case of a single clip, we reproduce the results of GSM, which is closest to our network architecture. 
Finally, we obtain 4\% increase in accuracy than GSM. 
In the end, we are 5.9\% better than Timeformer-L whose input clip contains 96 frames, achieving the best result so far. 
The experiments on Diving fully demonstrate the advantages of SGN in capturing long-term dependencies and modelling the global structure of video clips.

\textbf{Kinetics-400}. In general, for the kinetic-400 dataset, using scene information alone can already obtain considerable performance. 
We compare SGN to the same type of methods and present the results in Table.~\ref{kinetics_sota}. 
We can still achieve competitive results on Kinetics-400. 
In the same type of methods, that is, under the framework of 2D network with the temporal module, we are very close to the best performance.

\begin{table}[t]
\begin{center}
\begin{tabular}{|c|c|}
\hline
Method &Top-1(\%)\\
\hline\hline
SlowFast \cite{feichtenhofer2019slowfast} &77.6\\
TimeSformer \cite{2021Is} &74.9\\
TimeSformer-HR \cite{2021Is} &78.0\\
TimeSformer-L \cite{2021Is} &81.0\\
\hline 
GSM(our impl.) \cite{sudhakaran2020gate}      &80.7  \\
\hline  
SGM(ours)           &  \textbf{84.7}   \\
SGM(double-clips)(ours)            &\textbf{86.9}\\
\hline 
\end{tabular}
\end{center}
\caption{Comparison with state-of-the-art methods on diving48.}
\label{diving_sota}
\end{table}

\begin{table}[t]
\begin{center}
\begin{tabular}{|c|c|}
\hline
Method &Top-1(\%)\\
\hline\hline
3rd &35.9\\
2nd &37.0\\
\hline  
SGM           &  \textbf{37.9}   \\
SGM($en$)(1st)            &\textbf{45.4}\\
\hline 
\end{tabular}
\end{center}
\caption{Result of MMVRAC Fisheye Video-based Action Recognition competition (ICCV21).}
\label{uav}
\end{table}

\textbf{UCF101 and HMDB51}. UCF101 and HMDB51 are two small datasets, and we transfer the model pre-trained on Kinetics to them to test the generalization of the proposed SGN. 
We report the average performance in Table.~\ref{ucf_sota} over three splits with 16 frames as input. 
Since HMDB51 relies more on temporal cues, we achieve the best results on HMDB51 and acceptable results on UCF101.

\subsection{Visualization}
\begin{figure}
\begin{center}
\includegraphics[width=1\linewidth]{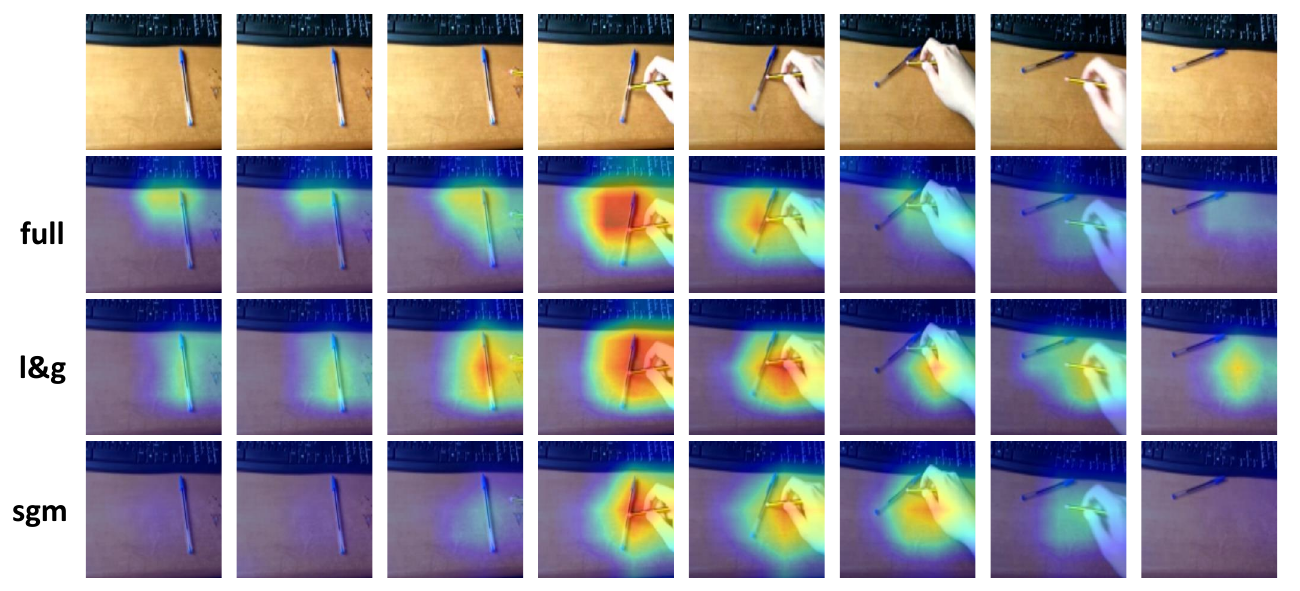}
\end{center}
\caption{Visualization of activation maps with Grad-CAM. The first row is the input video clip.}
\label{grad_cam}
\end{figure}

We first use Grad-CAM \cite{selvaraju2017grad} to visualize the class activation map. 
Fig.~\ref{grad_cam} shows the results. 
The results indicate that the model with a complete graph module ignores the keyframe of the action. 
When decomposing the graph into local and global subgraphs, the attention of the keyframes is increased, but some noise frames are also concerned. 
Finally, in SGM, the model only pays attention to the keyframes. 
The visualization results show that by splitting the complete graph into multiple subgraphs, the mixed temporal cues are gradually separated.

\begin{figure}
\begin{center}
\includegraphics[width=1\linewidth]{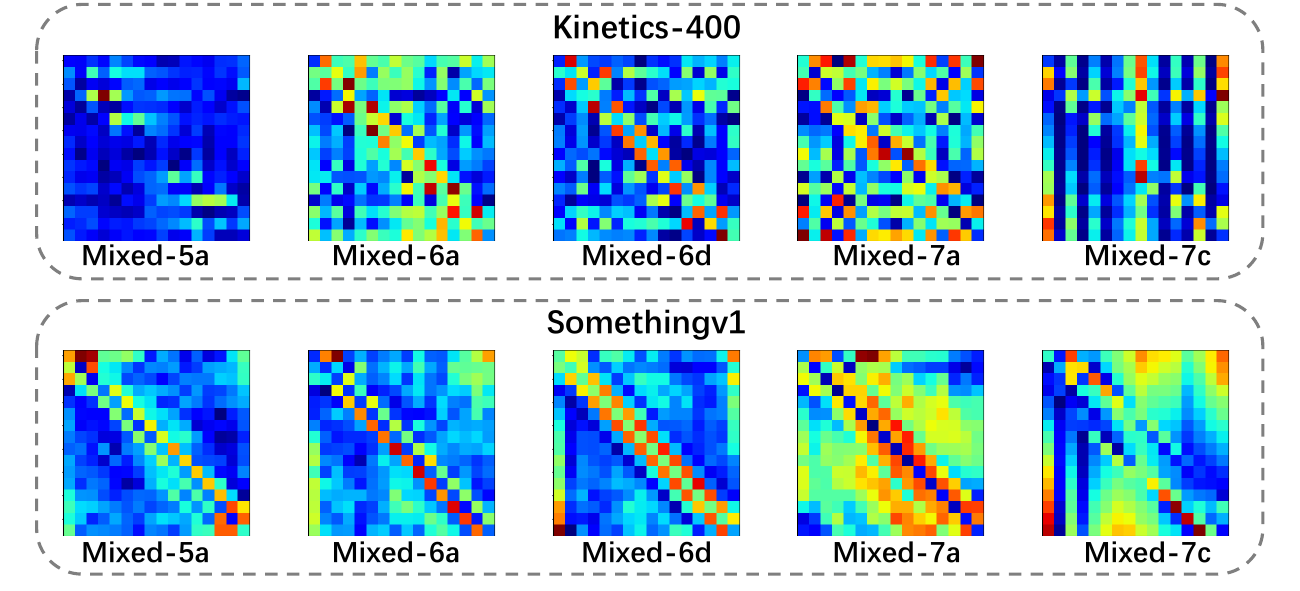}
\end{center}
\caption{Adjacent matrices in different SGN layers on Kinetics-400 and Something-somgthing V1. In a matrix, The vertical dimension represents temporal position and the horizontal dimension represents relationships. So the value at row $i$ column $j$ represents the weight of the edge from frame $j$ to frame $i$.}
\label{matrix}
\end{figure}

We also visualize the adjacent matrices learned in different layers of SGN. 
As shown in Fig.~\ref{matrix}, different temporal patterns are concerned in different layers. 
With the increase in depth, the global features are increasingly valued. 
And for different datasets, the adjacency matrix of the same layer is also different. 
For the Something-something V1 dataset with a shorter sample duration, the local features are emphasized, while in the longer Kinetics-400, the global information is paid more attention.

\subsection{MMVRAC Fisheye Video-based Action Recognition competition (ICCV2021)}
The SGN network was used in the 1st solution in MMVRAC Fisheye Video-based Action Recognition competition (ICCV21). 
In Table.~\ref{uav} we report the performance of our SGN and the top-3 solutions. 
The final 1st-place solution is an ensemble of multiple models trained by several training strategies. 

\section{Conclusion}
In this paper, We abandoned the popular way of viewing video clips as sequences and proposed to think of them as graphs with interconnected components. 
When transferring clips to graph representations, we notice the problem of sequence information loss in the transformation process and propose the structural decomposition idea (SGM) to alleviate this problem. 
We take the temporal prior attributes as the basis to guide the decomposition, so that SGM can capture the global structural information and sequence features of the video clips simultaneously. 
We designed sufficient ablation experiments to demonstrate the effectiveness of SGM. 
Finally, the promising results on 6 popular action recognition datasets suggest that our method can obtain SOTA performance at a relatively lower computation cost. 
In general, the most important contribution of our method is to treat the temporal distribution of video clips from a new perspective and put forward an effective practical scheme. We hope that the idea of viewing the temporal dimension of videos from a graph perspective will get more attention. 
\section*{Acknowledgement}
This work was supported by the National Natural Science Foundation of China (U1836218, 62020106012, 62106089), and the 111 Project of Ministry of Education of China (B12018).

\ifCLASSOPTIONcaptionsoff
  \newpage
\fi

\bibliographystyle{IEEEtran}
\bibliography{egbib}

\begin{IEEEbiography}[{\includegraphics[width=1in,height=1.25in,clip,keepaspectratio]{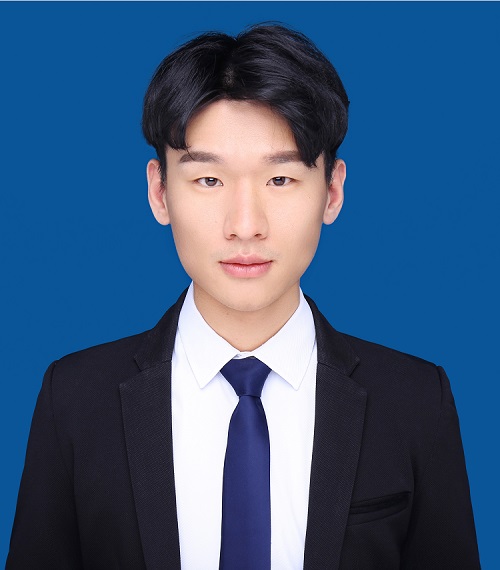}}]{Rongchang Li} received the B.Sc. degree from School of Civil Engineering, Tianjin Univercity, China, in 2019. He is currently pursuing the PhD degree with the Jiangsu Provincial Engineerinig Laboratory of Pattern Recognition and Computational Intelligence, Jiangnan University. His research interests include action recognitoin and video representation. He achieved top 1 recognizing performance on two tracks (Track-2 and Track-3) in Multi-Modal Video Reasoning and Analyzing Competition (ICCV21).
\end{IEEEbiography}

\begin{IEEEbiography}[{\includegraphics[width=1in,height=1.25in,clip,keepaspectratio]{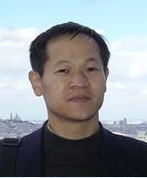}}]{Xiaojun Wu}received the B.Sc. degree in mathematics from Nanjing Normal University, Nanjing,
China, in 1991, and the M.S. degree and Ph.D. degree in pattern recognition and intelligent system from the Nanjing University of Science and Technology, Nanjing, in 1996 and 2002, respectively. From 1996 to 2006, he taught at the School of Electronics and Information, Jiangsu University of Science and Technology, where he was promoted to Professor. He has been with the School of Information Engineering, Jiangnan University since 2006, where he is a Professor of pattern recognition and computational intelligence. He was a Visiting Researcher with the Centre for Vision, Speech, and Signal Processing (CVSSP), University of Surrey, U.K. from 2003 to 2004. He has published over 300 papers in his fields of research. His current research interests include pattern recognition, computer vision, and computational intelligence. He was a Fellow of the International Institute for Software Technology, United Nations University, from 1999 to 2000. He was a recipient of the Most Outstanding Postgraduate Award from the Nanjing University of Science and Technology.
\end{IEEEbiography}

\begin{IEEEbiography}[{\includegraphics[width=1in,height=1.25in,clip,keepaspectratio]{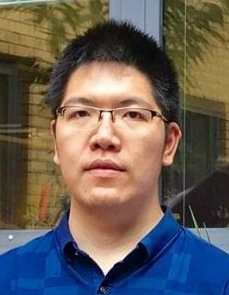}}]{Tianyang Xu} received the B.Sc. degree in electronic science and engineering from Nanjing University, Nanjing, China, in 2011. He received the PhD degree at the School of Artificial Intelligence and Computer Science, Jiangnan University, Wuxi, China, in 2019. 
He is currently an Associate Professor at the School of Artificial Intelligence and Computer Science, Jiangnan University, Wuxi, China.
His research interests include visual tracking and deep learning. 
He has published several scientific papers, including IJCV, ICCV, TIP, TIFS, TKDE, TMM, TCSVT etc. He achieved top 1 tracking performance in competitions, including the VOT2018 public dataset (ECCV18), VOT2020 RGBT challenge (ECCV20), Anti-UAV challenge (CVPR20), Multi-Modal Video Reasoning and Analyzing Competition (ICCV21).
\end{IEEEbiography}

\end{document}